\documentclass[runningheads]{llncs}
\usepackage[T1]{fontenc}
\usepackage{graphicx}
\usepackage{subcaption}
\usepackage{hyperref}
\usepackage{booktabs}
\usepackage{tablefootnote}
\usepackage{xcolor}
\usepackage{threeparttable}
\usepackage{graphicx}
\usepackage{multirow}
\usepackage{tabularx}
\usepackage{csquotes}

\begin{document}
%




\title{Tacit Knowledge Extraction via Logic Augmented Generation and Active Inference}
\titlerunning{Tacit Knowledge Extraction via LAG and Active Inference}

\author{Lorenzo Lamazzi\inst{1}\orcidID{0009-0004-9808-2936} \and
Aldo Gangemi\inst{1,2}\orcidID{0000-0001-5568-2684} \and 
Alessio Giberti\inst{3} \and 
Andrea Giovanni Nuzzolese\inst{1}\orcidID{0000-0003-2928-9496} \and
Vittorio Andrea Rocca\inst{4} \and
Mattia Torta\inst{4} \and
Francesco Poggi\inst{1}\orcidID{0000-0001-6577-5606}}
\authorrunning{L. Lamazzi et al.}
\institute{CNR - Institute of Cognitive Sciences and Technologies, Bologna, Italy \\
\email{\{lorenzolamazzi,andreagiovanni.nuzzolese@cnr.it,francesco.poggi\}@cnr.it} \and
Department of Philosophy, University of Bologna, Bologna, Italy
\\
\email{aldo.gangemi@unibo.it} \and
MISTER Smart Innovation, Bologna, Italy \and
Consorzio MUSP, Piacenza, Italy
}

\maketitle              
\begin{abstract}

Tacit knowledge plays a central role in human expertise, yet it remains difficult to capture, formalize, and reuse in machine-interpretable form. This challenge is especially relevant in procedural domains, where successful execution depends not only on explicit instructions, but also on implicit assumptions, contextual constraints, embodied skills, and experience-based judgments rarely documented. 
As a result, current knowledge engineering pipelines struggle to transform tacit and process-centric knowledge into formally specified, machine-interpretable representations that can be queried, validated, reasoned over, and reused.

In this paper, we introduce a neuro-symbolic framework that combines Logic-Augmented Generation and an Active-Inference-inspired approach for ontology-grounded Knowledge Graph construction.
We evaluate the approach in a knowledge transfer case study in manufacturing, using assembly-like repair procedures from instructional videos as a reproducible proxy domain. 
Results show that the proposed solution improves completeness and semantic quality, advancing neuro-symbolic knowledge engineering for industrial domains.

\end{abstract}

\section{Introduction}

Procedural knowledge concerns knowing how to perform an activity, in contrast
to knowing that certain facts hold~\cite{ryle1949concept}. In cognitive science,
this distinction is commonly related to the contrast between procedural
knowledge, which supports actions and skills, and declarative knowledge, which
represents facts and propositions~\cite{anderson1983architecture}. Procedural
knowledge is therefore central in domains where expertise is expressed through
goal-oriented sequences of actions, such as repairing a device, assembling a
component, operating a machine, or performing a maintenance procedure.

However, procedural knowledge is rarely exhausted by explicit instructions.
Experts often rely on perceptual, motor, contextual, and domain-specific
assumptions that remain unstated. For example, in a culinary recipe, the
instruction \enquote{\textit{whisk the eggs until stiff peaks form}} implicitly assumes that only
egg whites are used, that they are placed in a bowl, and that a manual or
electric whisk is used. Similarly, in repair or assembly, \enquote{\textit{tighten the screws}}
presupposes a screwdriver compatible with the screw head. These assumptions
correspond to tacit knowledge, classically described by Polanyi through the
observation that humans can know more than they can tell~\cite{polanyi1966tacit}.


In this paper, we treat tacit knowledge as a critical component of procedural knowledge. 
It includes both embodied know-how, such as force, timing, alignment, and perceptual judgment, and domain-presupposed knowledge, such as implicit tools, prerequisites, warnings, and contextual constraints. This distinction is central for knowledge transfer, i.e., the process through which knowledge held by individuals or groups is made available to others and affects their ability to perform tasks~\cite{argote2000knowledge}. A novice may follow the documented sequence of operations and still fail if relevant tacit conditions remain implicit.

To be usable for training, documentation, automation, or decision support, procedural knowledge must be transformed into formally specified, machine-interpretable representations that can be queried, validated, reasoned over, and reused. Ontologies and Knowledge Graphs provide a natural foundation for this objective, but current extraction pipelines face two main limitations. First, they mainly capture information that is explicitly stated or directly observable, producing graphs that may be structurally valid but procedurally incomplete because relevant tacit knowledge remains implicit. Second, when based on unconstrained LLM generation, they may introduce hallucinations, inconsistent schemas, unsupported entities, and poorly traceable inferences. 
%
%

To address these problems, we propose a neuro-symbolic framework that combines Logic-Augmented Generation (LAG)~\cite{gangemi2025} and an Active-Inference-inspired strategy~\cite{FRISTON2016862} for ontology-grounded Knowledge Graph (KG) construction. LAG constrains the Large Language Models (LLM) with a reference ontology and guides the generation of Resource Description Framework (RDF) assertions that instantiate the target schema. It extracts explicit procedural knowledge from instructional sources and represents it as a validated KG. 
The Active-Inference-inspired strategy then enriches the graph with tacit elements by prompting the model to reason over the source-grounded procedural structure and identify implicit elements, including tools, artifacts, assumptions, affordances, contextual constraints, and warnings.

We evaluate the proposed framework in a knowledge transfer case study in the manufacturing domain. Since large-scale, publicly available video datasets of industrial assembly processes are limited, we use assembly-like repair procedures from instructional videos as a reproducible proxy domain. These procedures involve fine-grained manipulation actions, specialized tools, ordered operations, component handling, and safety-critical constraints, making them suitable for
studying the extraction and enrichment of procedural knowledge. The evaluation benchmarks heterogeneous LLMs, including proprietary, open-source, and multimodal models, across transcript-based and vision-based settings.
The rest of the paper is organized as follows: Section~\ref{sec:related} discusses related work; Section~\ref{sec:architecture} presents the system
architecture, including the reference ontology, the LAG pipeline, and the Active-Inference-inspired enrichment strategy. Section~\ref{sec:experiments} describes the experimental setup; Section~\ref{sec:evaluation} reports and discusses evaluation results. Finally, Section~\ref{sec:conclusion} draws conclusions and outlines future work.


\section{Background and Related Work} \label{sec:related}

KG construction is a central task in Semantic Web research, aiming to 
represent entities, relations, and events from heterogeneous sources in a structured and semantically explicit form. Ontologies and graph schemas provide shared vocabularies, typing constraints, and semantics for querying, integration, validation, and reasoning~\cite{Hogan2022}. Automatic KG construction is commonly organized around knowledge acquisition, refinement, and evolution, where acquisition extracts structured knowledge and refinement improves completeness, consistency, and quality~\cite{zhong2024}. Refinement includes completion, error detection, and quality assessment, showing that coverage and correctness remain persistent challenges even for explicit factual knowledge~\cite{Paulheim2017489}. These issues become even more critical in procedural domains, where relevant knowledge includes not only entities and relations explicitly mentioned in a source, but also implicit tools, contextual constraints, and tacit assumptions required for execution.

Recent work has explored LLMs for Knowledge Engineering, including information extraction, ontology population, and ontology-guided KG generation. Text2KGBench~\cite{Mihindukulasooriya2023247}, presented at ISWC 2023, evaluates ontology-constrained KG generation from text while measuring ontology compliance and faithfulness to the input. However, it mainly targets source-grounded factual extraction, leaving open the problem of representing procedural and tacit knowledge conveyed through action sequences and multimodal evidence.

The integration of LLMs with symbolic representations is increasingly studied to mitigate hallucinations, improve faithfulness, and make generated outputs more inspectable. Logic-Augmented Generation (LAG) formulates this idea in the Semantic Web context by using KGs to introduce logical and factual boundaries for LLM-based generation~\cite{gangemi2025}. We build on this perspective by using a reference ontology as a generative constraint: the LLM must produce RDF assertions that instantiate the target schema, and the resulting graph is validated through ontology-based and SHACL constraints.

While ontology-constrained generation can improve structural validity and reduce unsupported schema-level outputs, it does not fully address the problem of tacit procedural knowledge. Active Inference offers a useful conceptual framework for reasoning from observations to hidden states, plausible policies, and action-relevant assumptions~\cite{FRISTON2016862,Friston20171}. In this paper, we do not implement Active Inference as a full cognitive architecture; rather, we use it as an inspiration for structuring the enrichment process. Starting from source-grounded procedural observations, the model is guided to infer implicit tools, artifacts, affordances, contextual constraints, warnings, and prior beliefs that make the observed procedure executable. Each inferred element is kept separate from explicit observations and associated with a justification and a confidence score.

Differently from prior work on ontology-driven KG generation, we address procedural KG construction from instructional videos and transcripts, with particular attention to tacit and domain-presupposed knowledge. The framework combines LAG for source-grounded extraction with Active-Inference-inspired enrichment, allowing the graph to represent both explicit observations and implicit tools, artifacts, assumptions, warnings, and contextual constraints required for execution.

\section{System Architecture} \label{sec:architecture} 
This section presents the system architecture of the proposed framework for extracting and enriching procedural knowledge from instructional videos. The architecture is organized around three main components: a reference ontology that defines the semantic model used to represent procedural knowledge; a Logic-Augmented Generation (LAG) pipeline that processes video-based or transcript-based input and generates ontology-grounded KGs; and an Active-Inference-inspired module that infers tacit procedural elements and represents them as controlled enrichments of the graph. The following subsections describe each component and explain how they interact within the overall workflow. 
Supplementary resources including the ontology, the implementation code for both the LAG pipeline and the Active Inference module, the two prompts used in the framework, and representative examples are available in the related GitHub repository
~\footnote{\url{https://github.com/LoLmz/tacit-knowledgex}}.

\subsection{Ontology Description}
\begin{figure*}[t]
    \centering
    \includegraphics[width=1\linewidth]{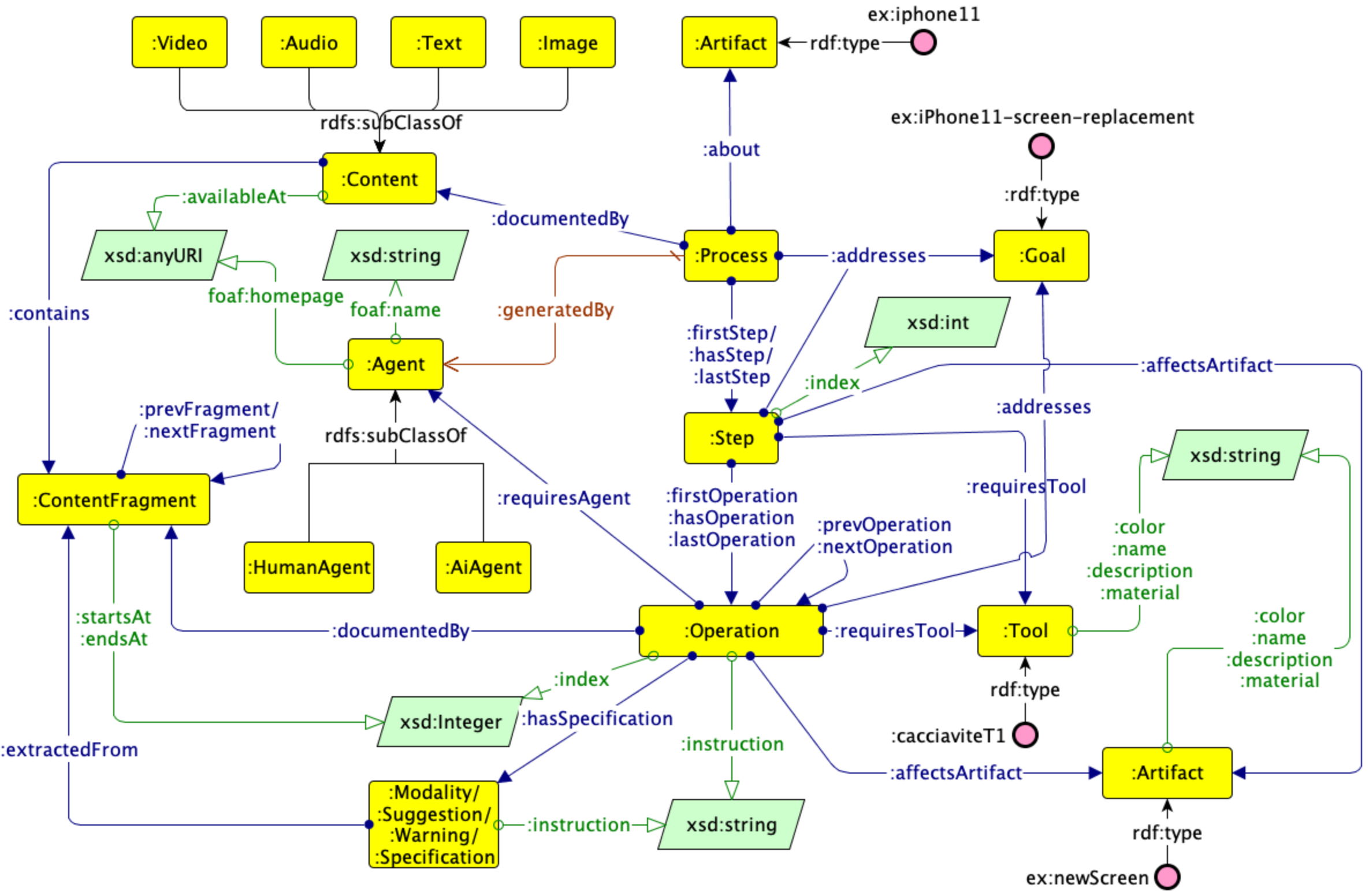}
    \caption{Overview of the proposed ontology.}
    \label{fig:ontology}
    \vspace{-0.5cm}
\end{figure*}

To support the structured representation of procedural knowledge, we define an ontology\footnote{The ontology is available in the GitHub repository associated with this work and will be made publicly accessible through relevant repositories after the review process.} that captures the main entities involved in a process. Figure~\ref{fig:ontology} provides an overview of the main classes and relations included in the ontology.
The ontology has been designed to support the semantic representation of procedural workflows, operational instructions, and multimodal documentation associated with repair processes. In particular, the ontology provides a conceptual framework for modelling processes as structured sequences of steps and operations performed on physical artifacts using tools, while preserving links to the original multimedia sources from which the knowledge is extracted.
The ontology was developed following an iterative ontology engineering approach based on eXtreme Design (XD)~\cite{Blomqvist2016}, thus inspired by the re-use of Ontology Design Patterns (ODPs)~\cite{Gangemi2005}. The design process focused on identifying the main entities and relations required to formally describe procedural knowledge such as textual manuals, videos, images, and audio recordings. The resulting conceptualization enables the representation of 
procedural execution and provenance information, thus supporting traceability between extracted knowledge and source content.

At the core of the ontology is the class \texttt{Process}, which represents a complete assembly, disassembly, or maintenance procedure. Following the design of the Sequence ODP\footnote{\url{http://ontologydesignpatterns.org/cp/owl/sequence.owl}.} a process is composed of an ordered sequence of \texttt{Step} instances connected through explicit sequential relations such as \texttt{nextStep}, \texttt{prevStep}, \texttt{firstStep}, and \texttt{lastStep}. Each step may contain one or more \texttt{Operation} instances representing atomic actions performed during the execution of the procedure. Operations may include actions such as unscrewing, disconnecting, prying, or cleaning.

The ontology explicitly models the artifacts affected during a procedure through the class \texttt{Artifact}. Operations and steps can be linked to artifacts via the property \texttt{affectsArtifact}, thereby enabling the representation of the components involved in each procedural action. Furthermore, tools and auxiliary materials required during the execution of a procedure are represented through the class \texttt{Tool}. This modelling choice allows the ontology to capture not only procedural actions but also the operational context in which they occur. An important aspect of the ontology concerns the representation of operational goals and specifications. The class \texttt{Goal} is used to represent the intended objective addressed by a process, step, or operation through the property \texttt{addresses}. In addition, the class \texttt{OperationSpecification} enables the formalisation of qualitative and quantitative constraints describing how an operation should be performed. Examples include torque values, material specifications, safety instructions, or qualitative criteria for completing a task. Operations can therefore be enriched with procedural constraints that are relevant for accurate execution.

Procedural knowledge may be documented by instances of the class \texttt{Content}, which abstracts over different media types including \texttt{Video}, \texttt{Image}, \texttt{Text}, and \texttt{Audio}, thus enabling the representation of multimodal content. Each content resource may contain one or more \texttt{ContentFragment} instances corresponding to delimited portions of the source material, such as temporal segments of a video or textual excerpts. Through properties such as \texttt{documentedBy}, \texttt{contains}, and \texttt{extractedFrom}, the ontology preserves explicit links between procedural entities and the original source fragments from which they have been derived. This feature is particularly important for explainability, traceability, and validation of extracted procedural knowledge.

The ontology further models the agents involved in generating or performing procedural knowledge. The class \texttt{Agent} abstracts over both human and artificial entities participating in the process, while subclasses such as \texttt{HumanAgent} and \texttt{AIAgent} distinguish between human operators and AI-based systems. Processes may therefore be associated with the entities responsible for generating, executing, or documenting procedural information.


\subsection{Logic Augmented Generation pipeline}
\begin{figure*}[t]
    \centering
    \includegraphics[width=1\linewidth]{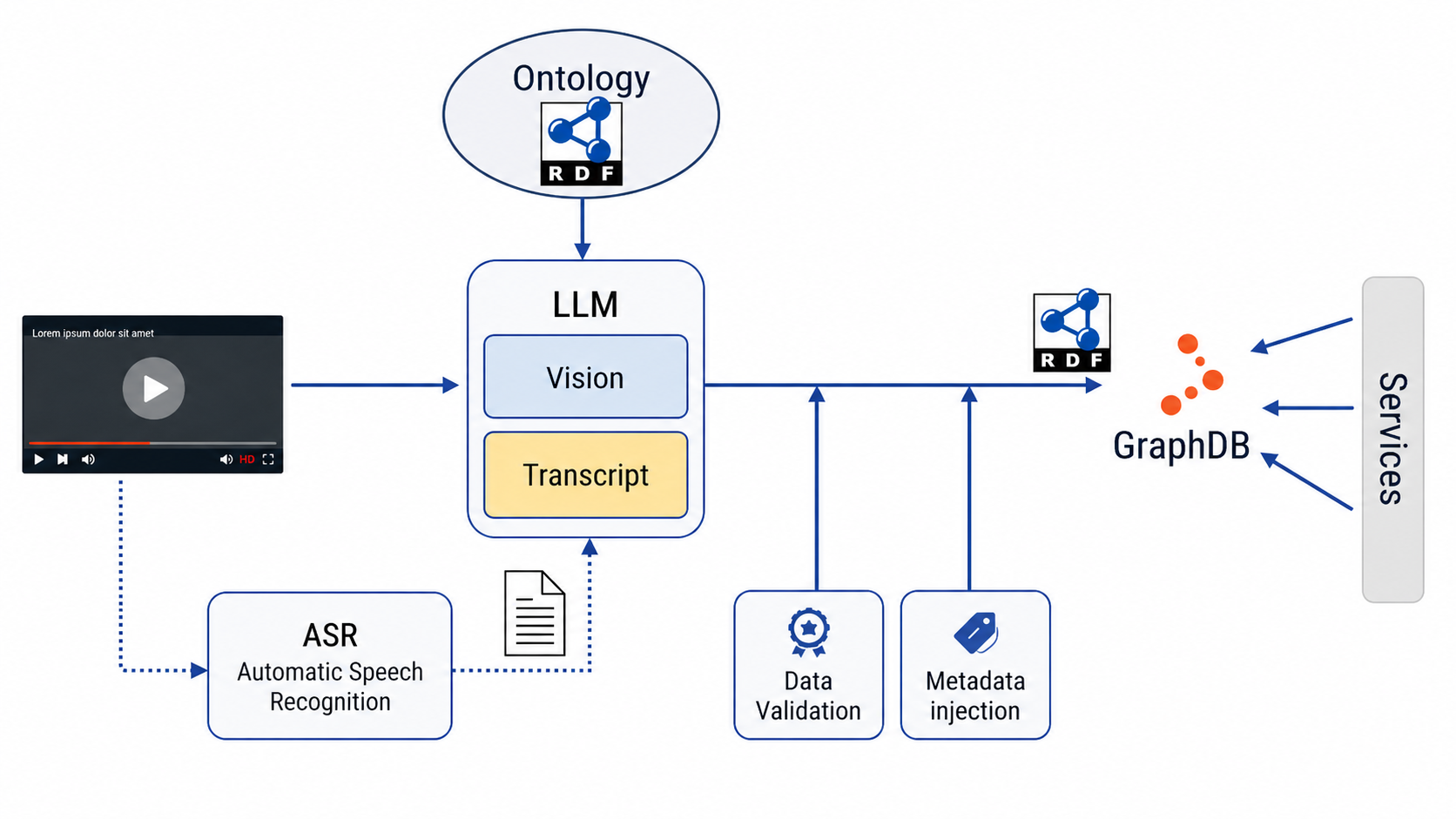}
    \caption{LAG pipeline.}
    \label{fig:pipeline}
    \vspace{-0.5cm}
\end{figure*}

The LAG approach is implemented through the pipeline illustrated in Figure \ref{fig:pipeline}. The pipeline shows how procedural knowledge is extracted from instructional videos, aligned with the reference ontology, validated, and represented as a KG.
The framework takes as input a reference ontology defining the target conceptual model, a source to be analyzed (i.e. an instructional video or its
transcript, depending on the capabilities of the selected model), and a prompt specifying the expected extraction task and output format.
These elements constrain the generation process by defining the vocabulary that the model can use to describe procedural entities, actions, tools, artifacts, and their relations.

The pipeline is model-agnostic and supports two input modalities depending on the capabilities of the selected LLM. If the model supports video input, the instructional video is passed directly alongside the ontology and the prompt, allowing the model to exploit both visual and linguistic content. If the model only supports textual input, the video is first transcribed using Whisper \cite{radford2023robust}, and the resulting ASR output is passed as text. In both cases, the prompt instructs the model to generate only individuals consistent with the reference ontology, without introducing new classes or properties.
The prompt was designed following general best-practice guidelines for structured output generation, drawing on the common recommendations shared across major model providers. \footnote{
Anthropic prompt engineering tutorial: \href{https://github.com/anthropics/prompt-eng-interactive-tutorial}{Github};
Google prompt strategies guide: \href{https://ai.google.dev/gemini-api/docs/prompting-strategies}{Guide};
OpenAI prompt engineering guide: \href{https://platform.openai.com/docs/guides/prompt-engineering}{Guide}.
}

The final output is a Turtle file instantiating the ontology and representing the extracted procedural knowledge as a KG.
The generated graph is then validated against the reference ontology through SHACL constraints. Shape-based validations verify that the extracted individuals comply with the expected ontological structure, while a set of global consistency checks detects major extraction failures — for instance, if the graph contains no instances of fundamental classes such as \texttt{Step} or \texttt{Artifact}, the process is interrupted and an error is reported. Only graphs that pass this validation stage are propagated to the subsequent components of the framework.
As a final step, the validated graph is loaded into a GraphDB triple store, where it is persistently stored and made available for SPARQL querying. The triple store acts as the central access point for analyzing, comparing, and exploiting the generated KGs across different experimental runs and downstream services.

Overall, the pipeline transforms procedural knowledge originally conveyed through instructional videos into a formal and machine-interpretable representation. Once stored in the triple store, this knowledge can be queried by different services that can present its content in multiple formats, such as a step-by-step web guide, a repair manual, or other user-oriented instructional resources.

\subsection{Active Inference Strategy}
Although the LAG pipeline guarantees source-grounded extraction, it can produce incomplete procedural representations, especially when the input is an ASR transcript and visual cues and implicit domain knowledge are unavailable.
To address this limitation, we introduce an Active-Inference-inspired enrichment strategy, extending the LAG pipeline with a dedicated reasoning prompt. In this work, the Active Inference module is evaluated only in transcript-based mode: therefore, even for models that process videos directly in the LAG pipeline, the enrichment strategy receives the ASR transcript as input.
Borrowing the notions of observation, hidden state inference, and policy from Active Inference theory, the prompt guides the LLM through a structured reasoning process that makes prior knowledge explicit and controllable, enabling the model to recover tacit operations, implicit tools, artifacts, warnings, and contextual constraints beyond what is directly verbalized in the input.
The reasoning process unfolds in four phases. In the observation phase, the model identifies the explicit procedural content available in the input: actions, mentioned entities, targets, and locations. In the hidden state inference phase, it reasons about the entities, tools, and environmental conditions that must be present for the observed actions to be physically possible. In the policy reconstruction phase, each step is decomposed into ordered sub-operations that preserve physical and procedural coherence. Finally, in the affordance reasoning phase, tools and artifacts are inferred from the actions they enable: if a procedure involves fastening, opening, aligning, or removing a component, the model infers which instruments are typically required.
For each inferred element, the model is asked to make explicit the prior belief supporting the inference — identifying the general repair pattern or procedural script that justifies it — and to assign a confidence score reflecting the reliability of the assumption. This allows the enriched graph to clearly distinguish directly observed procedural knowledge from model-inferred content, and provides a traceable justification for each tacit element, supporting human verification where needed.

\section{Experimental Setup} \label{sec:experiments}

This section describes the experimental setting used to evaluate the proposed framework, including the dataset, the models under evaluation, and the annotation protocol.

\subsection{Datasets and Scenarios}

In order to evaluate our framework under conditions that closely resemble the project’s real use case, we require a large dataset of video-based cases. However, the availability of publicly accessible video datasets in industrial manufacturing settings is limited, particularly given the level of detail required to analyse procedural knowledge, action sequences, tool usage and implicit decision patterns.
This scarcity makes it difficult to perform a systematic and reproducible assessment of real-world manufacturing footage. To address this, we have selected the iFixit video library as a proxy domain for our experiments. 
iFixit provides a large collection of repair videos for electronic and electromechanical devices, covering tasks that exhibit similarities with manufacturing and assembly processes. Although these videos are situated in a repair context rather than in a production environment, they involve structured sequences of fine-grained manipulation actions, the use of specialized tools, the identification and handling of components, and the execution of procedurally constrained operations. These characteristics make them particularly suitable for evaluating our framework in scenarios that approximate assembly industrial activities. 
We selected three videos covering different procedures on three distinct devices, as shown in Table~\ref{tab:resources}. The videos capture repair processes characterized by multiple sequential steps, intensive object and tool manipulation, and strict procedural constraints. Moreover, they provide fine-grained instructions that must be executed with precision to avoid damaging the device. Such characteristics make these scenarios particularly suitable for evaluating the ability of the proposed framework to extract explicit procedural knowledge and to infer implicit and tacit information. Overall, the resulting dataset enables the assessment of ontology-grounded knowledge extraction in conditions that are reasonably representative of the target domain, while still remaining manageable and reproducible for benchmarking purposes.

\begin{table}[t]
\centering
\caption{Video resources used in the evaluation.}
\label{tab:resources}
\scriptsize
\begin{tabular}{ll} 
\toprule
\textbf{TITLE} & \textbf{URL} \\
\midrule
iPhone 11 Battery Replacement
& \url{https://youtu.be/Feyqwf3cYtw?si=pusyAQIC99lcy2T7} \\
Display replacement Google Pixel 6a
& \url{https://youtu.be/EN2ibW2BWr0?si=lCaNWK80JbhzQLdQ} \\
Speaker replacement Sega Game Gear
& \url{https://youtu.be/6VFgXGt0KoY?si=3injiknTO5ueyIPk} \\
\bottomrule
\end{tabular}
\end{table}

\subsection{Models Under Evaluation}
As discussed in the previous sections, our evaluation considers two input modalities, namely a transcript-based setting and a vision-based setting. 
We benchmarked a heterogeneous set of LLMs including large proprietary models, a smaller proprietary model, and an open-source model. The selected models were chosen based on their popularity and practical relevance in current LLM-based workflows. Table~\ref{tab:models_under_evaluation} summarizes the main characteristics of the evaluated models.

The selected models cover the main comparison dimensions required by our evaluation. Gemini 3.1 Pro and Gemini 2.5 Flash are evaluated in both transcript-based and vision-based settings, enabling a comparison between text-only and video-native processing within the same model family. Claude Opus 4.6 and GPT-5.2 are included as proprietary transcript-based baselines, while Gemma 4 31b provides an open-source transcript-based baseline.

All models were tested using the same prompt structure and, within each modality, the same input content. Transcript-based models received the same ASR transcript, whereas vision-based models received the same instructional video. The vision-based setting was restricted to Gemini models, as they were the only models in our setup supporting direct video processing. This
configuration allows us to assess how model class, size, and modality affect ontology-grounded procedural knowledge extraction.

\begin{table}[t]
\centering
\caption{Models under evaluation}
\label{tab:models_under_evaluation}
\small


\begin{threeparttable}
\resizebox{\linewidth}{!}{%

\setlength{\tabcolsep}{5pt}
\begin{tabular}{lccrrrr}
\hline
\multicolumn{1}{c}{\multirow{2}{*}{\textbf{MODEL}}} & \multicolumn{1}{c}{\multirow{2}{*}{\textbf{OPEN}}} & \multirow{2}{*}{\textbf{MODALITIES}} & \multicolumn{1}{c}{\multirow{2}{*}{\textbf{\begin{tabular}[c]{@{}c@{}}CONTEXT\\ (tokens)\end{tabular}}}} & \multicolumn{1}{c}{\multirow{2}{*}{\textbf{\begin{tabular}[c]{@{}c@{}}MAX OUTPUT\\ (tokens)\end{tabular}}}} & \multicolumn{2}{c}{\textbf{PRICE (\$/1M)}} \\ \cline{6-7} 
\multicolumn{1}{c}{} & \multicolumn{1}{c}{} &  & \multicolumn{1}{c}{} & \multicolumn{1}{c}{} & \multicolumn{1}{c}{\textbf{in}} & \multicolumn{1}{c}{\textbf{out}} \\ \hline
Gemini 3.1 Pro & No & T, V & 1,048,576 & 65,536 & 2.00 & 12.00\tnote{b} \\
Gemini 2.5 Flash & No & T, V & 1,048,576 & 65,536 & 0.30 & 2.50 \\
Claude Opus 4.6 & No & T & 1,000,000 & 128,000 & 5.00 & 25.00 \\
GPT-5.2 Chat & No & T & 400,000 & 128,000 & 1.75 & 14.00 \\
Gemma 4 31b-it & Yes & T & 256,000 & npd & 0 & 0 \\ \hline
\end{tabular}

}

\begin{tablenotes}
\footnotesize
\item T = transcript-based setting; V = vision-based setting.
\item[b] Pricing reported for prompts up to 200k tokens.
\item npd = not publicly disclosed. 
\end{tablenotes}
\end{threeparttable}
\end{table}


\subsection{Experimental Protocol} 

To evaluate the quality of the extracted procedural knowledge, we established a manual annotation of the dataset as a reference baseline. A pool of domain experts manually identified the individual operations composing each procedure and annotated the tools and artifacts involved, resulting in more than 90 annotated operations across the analyzed videos.
The automatically extracted KGs were then compared against this reference annotation. Each experiment was repeated five times to assess output stability. Across runs, the generated KGs were highly consistent in terms of graph structure, recognized tools and artifacts, and absence of hallucinated entities. This indicates that the reported results are stable and mainly reflect differences in model behavior and input modality.
For tools and artifacts, the comparison was performed by matching the entities extracted by each LLM against the manually annotated ones, computing true positives (TP), false positives (FP), and false negatives (FN) at the operation level. These counts were used to derive precision and recall scores, providing a quantitative assessment of the model's ability to identify relevant procedural entities.
For the decomposition of procedures into steps and operations, and for the content of individual operation instructions, a direct quantitative evaluation based on precision and recall is not straightforward, as these elements do not admit a unique correct formulation. The assessment of their quality and practical usefulness is therefore deferred to future work, where the extracted procedural knowledge will be validated through a user study measuring its effectiveness in supporting real operators during task execution.

\section{Evaluation} \label{sec:evaluation}
This section reports the evaluation results, first discussing the LAG-based extraction and then the impact of the Active-Inference-inspired enrichment.

\subsection{LAG Results}


\subsubsection{Ontology Compliance}

A first relevant result concerns the syntactic and ontological validity of the generated outputs. Across all evaluated models and scenarios, the LAG pipeline produced valid Turtle files that could be loaded into the triple store without manual correction. Beyond syntactic validity, all generated graphs were fully compliant with the reference ontology: no model introduced individuals belonging to undefined classes or used properties outside the target schema. This result confirms that the LAG strategy effectively constrains the generative behavior of LLMs within the desired semantic model, reducing the risk of structurally inconsistent outputs at the schema level.
It is worth noting that ontology compliance does not imply semantic completeness. A model may still omit relevant entities or produce less informative descriptions while remaining structurally valid. This distinction motivates the more fine-grained evaluation presented in the following subsections.

\subsubsection{Step and Operation Decomposition}
The decomposition of procedures into steps and operations shows clear differences across the evaluated models. Although all models were able to represent the procedural structure of the source material, the level of granularity of the decomposition varied depending on the model and on the input modality.

Vision-capable models generally produced more accurate and complete decompositions as they could exploit visual information that was not explicitly verbalised in the transcript, thereby recovering actions that were shown but only partially described or entirely omitted in the narration. By contrast, transcript-based models were constrained by the linguistic content and produced decompositions that closely reflected the spoken narration. This resulted in a lack of completeness when relevant actions were conveyed through the visual channel.
Among the evaluated models, GPT-5.2 produced the least satisfactory decompositions, tending to group multiple heterogeneous actions into single operations rather than breaking the procedure down into individual steps. This reduces procedural granularity, making the resulting representation less suitable for downstream tasks requiring fine-grained action sequencing.

\subsubsection{Instruction Quality}
All of the models that were evaluated produced instructions that were procedurally meaningful, preserving the action-oriented nature of the source material and maintaining a clear relationship between the action, the affected component and its procedural context. Notably, all models consistently filtered out colloquial expressions, informal warnings and conversational remarks typical of instructional video narration, but which would be inappropriate in a structured KG. For example, phrases such as \enquote{you don't want to make this a two-part repair job} were correctly omitted or reformulated as explicit precautions. There were qualitative differences in the level of detail: some models generated concise instructions focused on the core action and its direct object, while others produced richer descriptions including contextual conditions and execution constraints.
 
Evaluating instruction quality in a rigorous and quantitative way is inherently difficult, as there is no single correct formulation for a procedural instruction, and correctness alone does not capture whether a description is sufficiently clear and actionable for a real operator. A meaningful assessment would require a user study in which participants attempt to execute the procedure using only the extracted instructions. This would involve measuring task success, error rate, and perceived clarity. This evaluation is therefore deferred to future work.

\subsubsection{Tool and Artifact Identification}
Tools and artifacts are fundamental in assembly and repair processes and provide a suitable category for quantitative evaluation, since they can be treated as discrete entities and compared against a reference annotation.
We therefore evaluate their extraction using precision, recall, and F1 score. Precision measures the proportion of extracted entities that are correct, recall measures the proportion of relevant entities that are successfully identified, and F1 summarizes the trade-off between the two.

\begin{figure*}[t]
    \centering
    \includegraphics[width=1\linewidth]{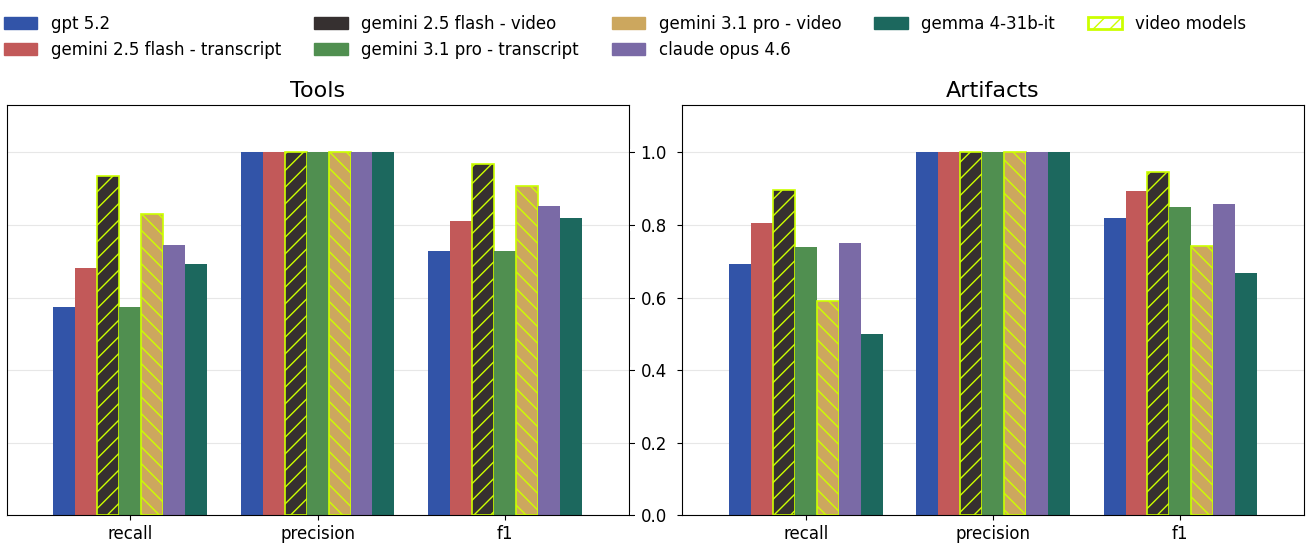}
    \caption{\textbf{LAG framework performance} on tools and artifacts extraction. Precision, recall, and F1 score are reported for all evaluated models. Transcript-based models are shown with solid bars, video-based models with highlighted hatched bars. All models achieve perfect precision (1.00), while recall varies substantially across models and modalities, with transcript-based models showing the largest gap.}
    \label{fig:lag_results}
    \vspace{-0.5cm}
\end{figure*}

Figure \ref{fig:lag_results} reports the results obtained by all evaluated models under the LAG framework. Precision is perfect (1.00) across all models and for both entity types. This result indicates that, under the LAG setting, models do not tend to hallucinate object-level entities: when a tool or artifact is extracted, it is grounded in the procedural source. This finding supports the role of LAG as a structural constraint that effectively suppresses the generation of unsupported elements in the KG.

Recall values, by contrast, show greater variability across models and modalities. Transcript-based models obtain recall scores ranging from 0.57 to 0.81 for tools and from 0.50 to 0.81 for artifacts. Among these, Gemma 4 31b-it, the only open-weight model evaluated, achieves competitive tool recall (0.69) but the lowest artifact recall observed across the entire evaluation (0.50, F1 = 0.67), suggesting that smaller open-weight models may struggle more with implicit or context-dependent artifact identification when visual information is unavailable. Video-based models generally perform better, reaching up to 0.94 for tools and 0.90 for artifacts with Gemini 2.5 Flash. However, access to video input does not always guarantee complete extraction, in fact Gemini 3.1 Pro with video input obtains a recall of only 0.59 for artifacts. These results confirm that the main limitation of the LAG framework is not the introduction of incorrect entities, but the omission of relevant ones, especially when they are implicit or conveyed through the visual channel.

\subsection{Active Inference Impact}
\begin{figure}[t]
    \centering
    \includegraphics[width=0.95\textwidth]{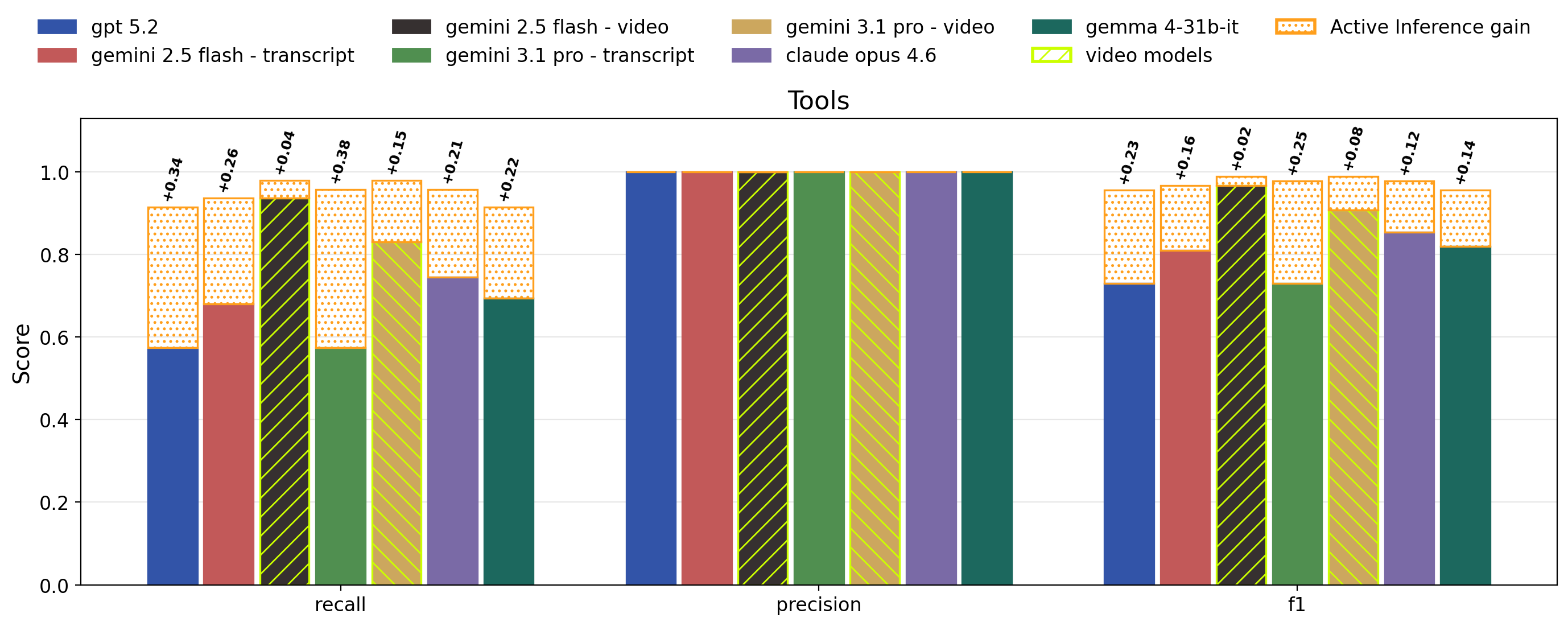}

    \vspace{-0.4em}

    \includegraphics[width=0.95\textwidth]{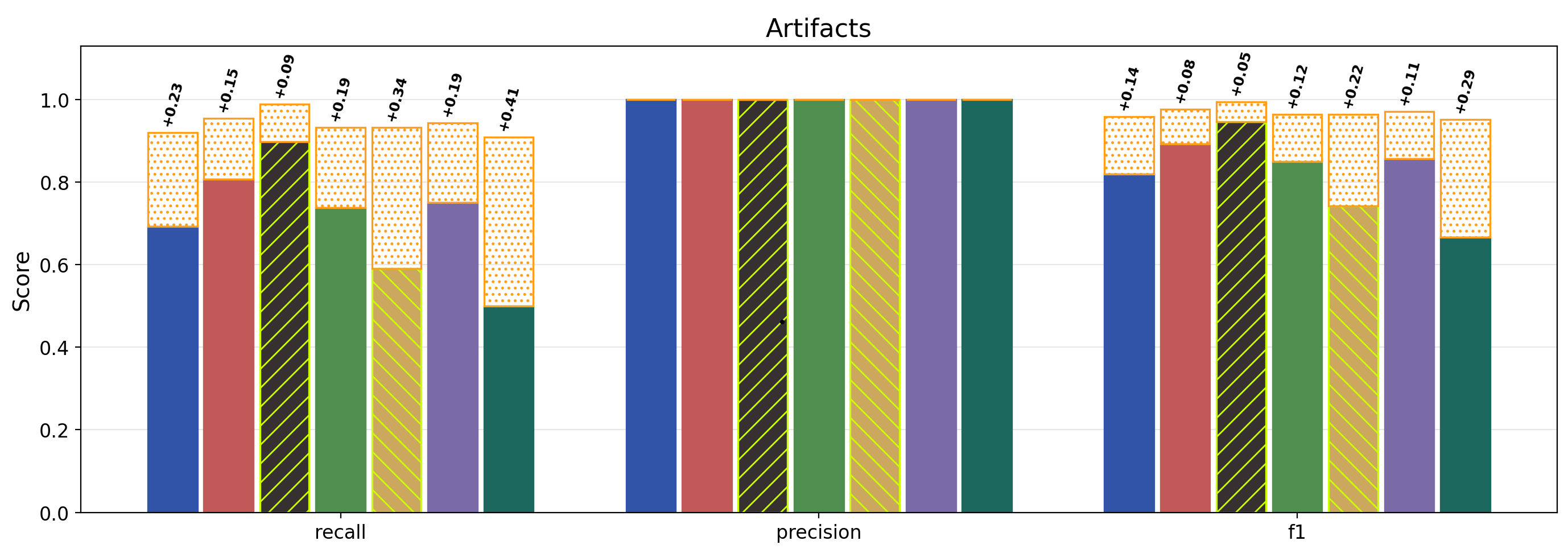}

    \caption{Active Inference impact on tool and artifact extraction. The upper and lower charts report results for tools and artifacts, respectively. 
    Precision, recall, and F1 score are reported before (solid/hatched bars) and after (orange dotted overlay) applying Active Inference. All models retain perfect precision and improve recall, with the largest recall gains observed for
transcript-based models.}
    \label{fig:active-inference-impact}
    \vspace{-0.8em}
\end{figure}

While LAG effectively produces ontology-compliant KGs and prevents schema-level false positives, it may still omit tools, artifacts, or operational details that are implicit, visually demonstrated but not verbalized, or assumed as common repair knowledge. To address this limitation, we applied the proposed Active Inference-inspired methodology. 
Figure \ref{fig:active-inference-impact} reports the impact of Active Inference methodology across all models, measured in terms of recall, precision, and F1 score for both tools and artifact extraction.


Since precision is already perfect across all models under the LAG framework, Active Inference contributes primarily by improving recall and, consequently, F1 score. This effect is most evident for transcript-based models, where the LAG stage suffers more from missing entities in the input source. For tools, recall increases range from +0.21 to +0.38 for transcript-based models, corresponding to F1 gains between +0.11 and +0.25. For artifacts, transcript-based models show recall gains between +0.19 and +0.41, with F1 improvements between +0.08 and +0.29. Video-based models generally show more modest gains, since they already achieved stronger baseline results with LAG, although Gemini 3.1 Pro with video input represents an exception for artifacts, improving by +0.34 in recall and +0.22 in F1. Overall, these results show that Active Inference is especially useful when relevant procedural entities are implicit or missing from the input modality. After enrichment, all models reach an F1 score above 0.95 for both tools and artifacts, with Gemini 2.5 Flash in video-based mode achieving the highest scores (0.989 for tools and 0.994 for artifacts).


This behavior is illustrated by a representative example. The transcript instruction \enquote{\textit{Lay the bracket that covers the connectors back on}} contains no reference to any tool. Consequently, the LAG pipeline correctly assigns no tool to this operation, as none is explicitly mentioned. Active Inference, however, infers that tweezers are implicitly required, classifying them as a tacit tool with the following justification: \enquote{\textit{Small brackets and screws are typically manipulated with tweezers to position and start screws safely.}} This inference is consistent with the visual evidence in the source video, where the technician is observed using tweezers, despite never verbalizing their use. The example illustrates how Active Inference recovers procedural knowledge that is demonstrated but not narrated, and therefore inaccessible to transcript-based extraction alone.


Beyond tool and artifact recovery, Active Inference also enriches the description of operations by making implicit accessory instructions and warnings explicit.
These elements are often fundamental for conveying how an operation should be performed, even when they are not directly stated in the source. 
For instance, the instruction \enquote{\textit{Align the bracket so that its screw holes line up with the threaded standoffs on the frame}} was inferred from a transcript that simply described placing the bracket back, without any reference to alignment. 
While this level of detail may appear self-evident to an experienced technician, it represents valuable procedural knowledge for less experienced operators.

\begin{table}[]
\centering
\resizebox{\linewidth}{!}{
\setlength{\tabcolsep}{4pt} 

\begin{tabular}{l|cc|cc|cccc}
\hline
\multicolumn{1}{c|}{\multirow{2}{*}{\textbf{Model}}} & \multicolumn{2}{c|}{\textbf{LAG extraction}} & \multicolumn{2}{c|}{\textbf{Act.Inf. enrichment}} & \multicolumn{4}{c}{\textbf{Full pipeline}} \\
\multicolumn{1}{c|}{} & \textbf{(\$/min)} & \textbf{(\$/hour)} & \textbf{(\$/min)} & \textbf{(\$/hour)} & \textbf{(\$/min)} & \textbf{(\$/hour)} & \textbf{\begin{tabular}[c]{@{}l@{}}Recall\\ Tools\end{tabular}} & \multicolumn{1}{c}{\textbf{\begin{tabular}[c]{@{}c@{}}Recall\\ Artifacts\end{tabular}}} \\ \hline
Gemini 3.1 Pro & 0.26 & 15.75 & 0.15 & 9.00 & 0.41 & 24.75 & 0.979 & 0.932 \\
Gemini 2.5 Flash & 0.06 & 3.30 & 0.03 & 1.88 & 0.09 & 5.18 & 0.979 & 0.989 \\
Claude Opus 4.6 & 0.08 & 4.73 & 0.31 & 18.75 & 0.39 & 23.48 & 0.957 & 0.943 \\
GPT-5.2 Chat & 0.04 & 2.63 & 0.18 & 10.50 & 0.22 & 13.13 & 0.915 & 0.920 \\
Gemma 4 31b-it & 0.00 & 0.00 & 0.00 & 0.00 & 0.00 & 0.00 &  0.915 & 0.909 \\ \hline
\end{tabular}

}
\caption{Cost and final recall of the evaluated models. Costs are reported for the LAG extraction stage, the Active-Inference enrichment stage, and the full pipeline. Recall values refer to the final KG after Active-Inference enrichment.}
\label{tab:costs}
\end{table}

\section{Conclusions and Future Work} \label{sec:conclusion}


\begin{figure}[t]
    \centering
    \begin{tabular}{cc}
        \includegraphics[width=0.49\textwidth]{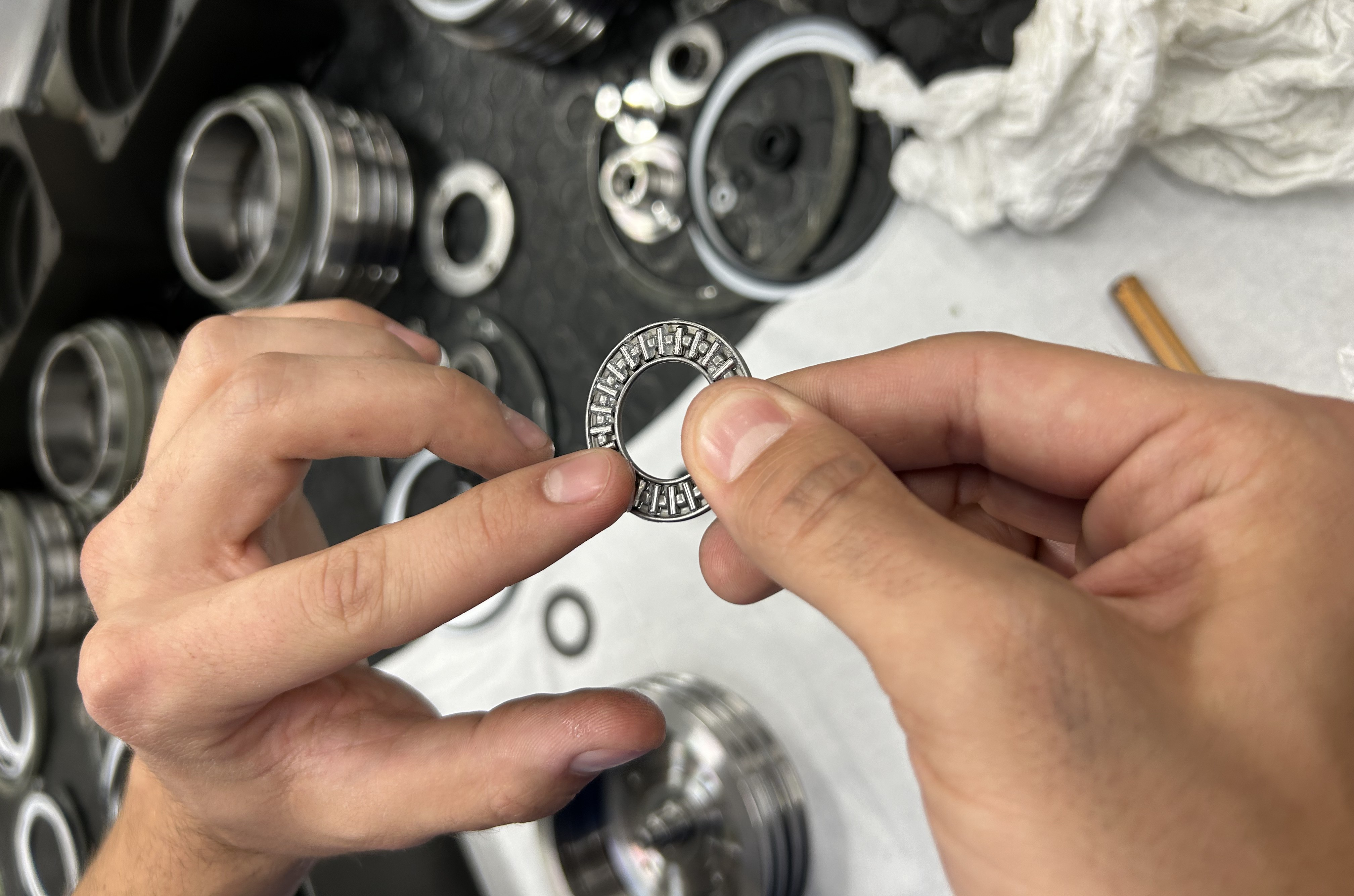} &
        \includegraphics[width=0.49\textwidth]{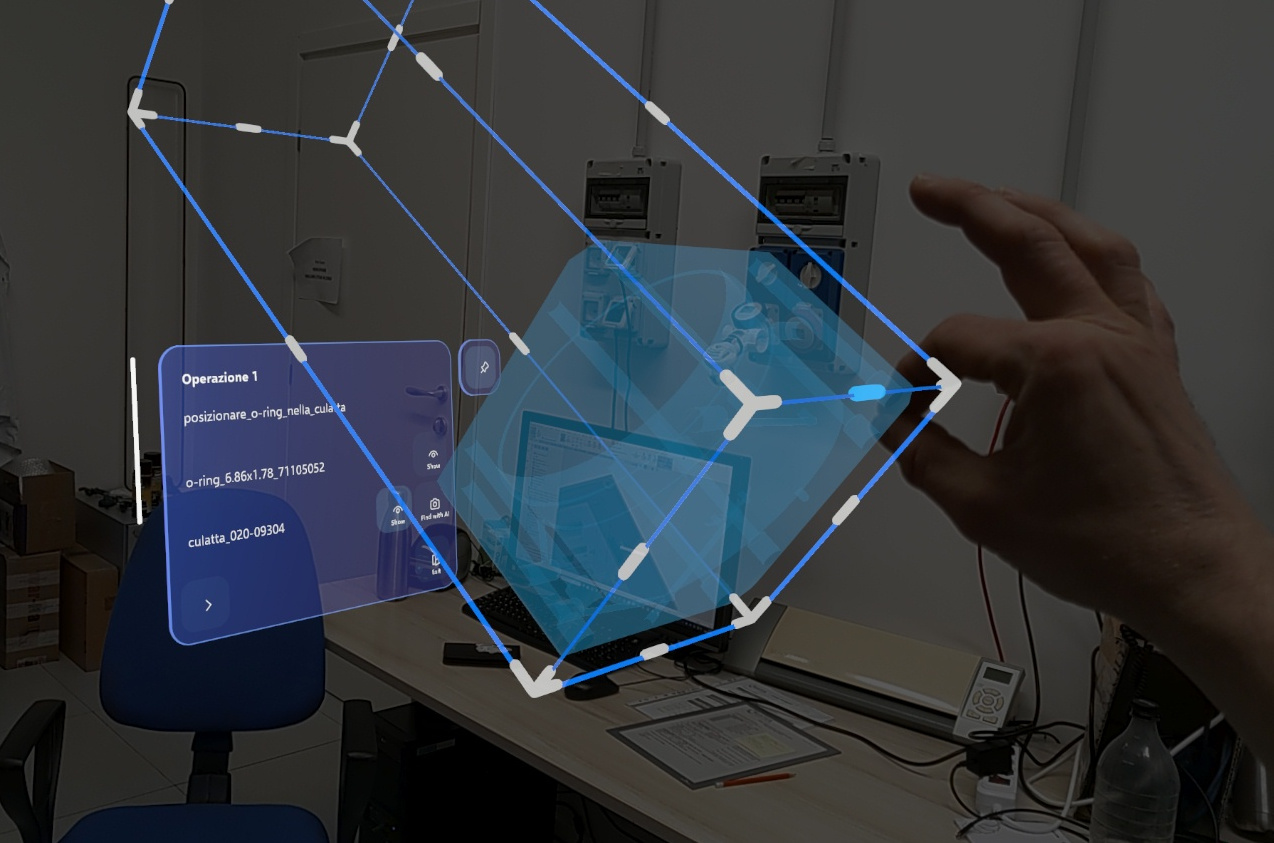} \\
        (a) & (b)
    \end{tabular}
    \caption{Application of the proposed framework. (a) Electrospindle assembly by an expert operator. (b) Augmented-reality environment for accessing procedural knowledge formalized as a KG.}
    \label{fig:kx}
\end{figure}

This paper presented a neuro-symbolic framework that combines LAG and an Active-Inference-inspired strategy for ontology-grounded procedural KG construction. By constraining LLM generation through a reference ontology and enriching the resulting graphs with explicitly represented tacit elements, the approach achieves strong results in terms of completeness and semantic quality, supporting the formalization of explicit and tacit procedural knowledge. 
The best results were obtained by Gemini 2.5 Flash in video-based mode, with an F1 score of 0.989 for tools, 0.994 for artifact, while Claude Opus 4.6 was the best transcript-based model, with an F1 score of 0.978 for tools and 0.971 for artifacts. The Active-Inference-inspired strategy was fundamental to improve completeness: recall increased on average by 0.30 for tools and 0.17 for artifacts in text-only models, and by 0.10 for tools and 0.22 for artifacts in multimodal models.

The cost analysis in Table~\ref{tab:costs} further supports the practical feasibility of the approach. Notably, Gemini 2.5 Flash is the least expensive proprietary configuration, with a full-pipeline cost of 5.18~\$/hour (0.09~\$/min), while also achieving the best final performance, with recall values of 0.979 for tools and 0.989 for artifacts. 
Larger proprietary models are substantially more expensive without yielding better results. The open-weight Gemma 4 31b-it obtains lower but still strong recall values, 0.915 for tools and 0.909 for artifacts, while requiring no API fee in our setup. This suggests that open models may be viable in contexts where cost, data governance, or local deployment are more important than maximizing extraction quality. Overall, the results indicate that the proposed framework can be used in real-world knowledge transfer scenarios with reasonable costs and high extraction quality.

The approach is currently being applied in 
the KnowledgeX project to real manufacturing knowledge transfer scenarios. Figure~\ref{fig:kx}(a) shows a frame from the assembly of an electrospindle by an expert operator, while Figure~\ref{fig:kx}(b) shows the augmented-reality environment developed to access and transmit the KG-based representation.

Future work will evaluate the approach on larger datasets and across different procedural domains. 
We also plan to investigate Active-Inference-based
techniques for detecting, while an expert operator performs the procedure to be transferred, tacit knowledge and other hot points that should be made explicit. This would allow the system to ask the expert operator in real time to provide further details and clarifications. Finally, we will study how the resulting KGs can support the semi-automatic generation of different knowledge access modalities, from instructions and manuals to interactive guidance and augmented-reality-based training environments.

\begin{credits}
\subsubsection{\ackname} This work was partially supported by the Emilia-Romagna Regional Program PR FESR 2021--2027 (Pr. KnowledgeX, grant no. E37G22000450007).

\end{credits}
%
%
%
\bibliographystyle{splncs04}
\bibliography{bibliography}
\end{document}